\documentclass[12pt,a4paper]{article}
\usepackage[top=3cm,bottom=2.5cm,left=2cm,right=2cm]{geometry}
\usepackage{mathtools}
\usepackage{float}
\usepackage{textcomp}
\usepackage[table]{xcolor}
\usepackage{graphicx,subfigure}
\usepackage{nth}
\usepackage{threeparttable}

\title{Joint Sentiment/Topic Modeling on Text Data Using Boosted Restricted Boltzmann Machine}
\author{Masoud~Fatemi, and~Mehran~Safayani
\thanks{M. Fatemi is with the Department of Electrical and Computer Engineering, Isfahan University of Technology, Isfahan 84156-83111, Iran e-mail: (m.fatemi@ec.iut.ac.ir).}
\thanks{M. Safayani is with the Department of Electrical and Computer Engineering, Isfahan University of Technology, Isfahan 84156-83111, Iran e-mail: (safayani@cc.iut.ac.ir, (corresponding author)).}}

\begin{document}
\maketitle

\noindent\textbf{Abstract}\\
\begin{small}
Recently by the development of the Internet and the Web, different types of social media such as web blogs become an immense source of text data. Through the processing of these data, it is possible to discover practical information about different topics, individual’s opinions and a thorough understanding of the society. Therefore, applying models which can automatically extract the subjective information from the documents would be efficient and helpful. Topic modeling methods, also sentiment analysis are the most raised topics in the natural language processing and text mining fields. In this paper a new structure for joint sentiment-topic modeling based on Restricted Boltzmann Machine (RBM) which is a type of neural networks is proposed. By modifying the structure of RBM as well as appending a layer which is analogous to sentiment of text data to it, we propose a generative structure for joint sentiment topic modeling based on neutral networks. The proposed method is supervised and trained by the Contrastive Divergence algorithm. The new attached layer in the proposed model is a layer with the multinomial probability distribution which can be used in text data sentiment classification or any other supervised application. The proposed model is compared with existing models in the experiments such as evaluating as a generative model, sentiment classification, information retrieval and the corresponding results demonstrate the efficiency of the method.\\
\end{small}

\section{Introduction}
\label{sec1}
Nowadays the ultimate objective of the artificial intelligence is to provide a way to perform different activities of human automatically and as quickly as possible. With the rapid expansion of the Internet in recent decades, various types of social media have been transformed into the massive sources of data, and especially text data, which can be processed to obtain valuable information about people\textquotesingle s viewpoints and general understanding toward different topics \cite{lin2012weakly}. The developments in the fields of text data mining and Natural Language Processing (NLP) have made major contributions to the understanding and analysis of these massive volumes of text data. Nevertheless, there is still a high demand for the methods capable of automatic analysis of massive unstructured data so as to extract the valuable information. Topic models are a class of text analysis methods that have recently drawn major attention from researchers of different fields, especially those working on NLP and text data mining \cite{mohr2013introduction}.

Topic models consider text documents as a mixture of multiple topics, where each topic can be treated as a probability distribution over words \cite{steyvers2007probabilistic}\cite{blei2012probabilistic}.
In the field of text data mining, topic models are those model that can detect and extract an abstract of the topics discussed in one or multiple documents \cite{blei2012probabilistic}\cite{blei2003latent}.
In other words, topic modeling methods are the tools that allow us to model text documents or any other set of discrete data. The goal of these models is to find a short description of the members of the dataset for effective processing of the original dataset without losing the statistical dependencies necessary for basic tasks such as classification or summarization \cite{blei2003latent}.

In the fields of NLP and text data mining, having a proper perspective about people\textquotesingle s manner of thinking is an important part of data collection \cite{pang2002thumbs}\cite{pang2008opinion}.
The advent and popularization of instant commenting tools such as online review sites and personal blogs has created new opportunities as well as challenges for understanding the people’s opinions through information technology \cite{pang2008opinion}.
In the field of opinion mining and sentiment analysis, the goal is to design and use a method or tool for automatic identification of conceptual information such as views, attitudes, and sentiments in a text document \cite{lin2012weakly}\cite{pang2002thumbs}.

While there have been many invaluable researches with major contributions to the above applications, there is a common deficiency in the existing body of literature and that is the concentration on detection of overall sentiment of documents without an in-depth analysis to identify latent topics and their associated sentiments. Each review contains and addresses a set of topics \cite{lin2012weakly}. For example, a review of a restaurant must address topics such as food, service, location, and price, among others. Detection of these topics is a necessary part of the process of retrieving more detailed information, but the absence of a sentiment analysis on the extracted topics may undermine the quality of the results. This is because users are interested not only in the overall sentiment of a review and its topical information, but also to the sentiment or opinion associated with each topic. For example, a customer may be content about the quality and price of food, but not about the service and location. Therefore, simultaneous detection of topics and their associated sentiments (join sentiment/topic modeling) is far more desirable as it provides information that is far more
valuable \cite{lin2012weakly}. In addition, detecting the sentiment of documents and topics can be as instrumental in the information retrieval as it is in topic detection in text mining. Hence, the methods of automatic joint sentiment/topic modeling of text documents could be of great scientific and economic value.

The present paper is focused on the processing of text data. Our goal is to use the capabilities of artificial neural networks to determine the distribution of topics discussed in the documents of a database and the word distribution and sentiments associated with each topic. In the text mining literature, the aforementioned process is known as joint sentiment/topic modeling. The good performance of neural networks in topic modeling, especially when compared to previous approaches that use Bayesian structures \cite{larochelle2012neural}\cite{hinton2009replicated}, and the limitations of Bayesian methods such as practical impossibility of exact inference \cite{hinton2009replicated}, necessitate more attention to the potentials and use of neural networks in this application. The proposed approach is a supervised generative probabilistic model based on the Restricted Boltzmann Machine (RBM) neural network \cite{smolensky1986information}\cite{hinton2002training} for the joint sentiment/topic modeling of text data.
Like other RBM-based methods, the model is trained using the Contrastive Divergence (CD) \cite{hinton2002training}\cite{woodfordnotes}\cite{hinton2010practical}\cite{carreira2005contrastive} learning algorithm.

The rest of this paper is organized as follows: In the second section, we review the previous work on the estimation of probability distributions in input data, topic modeling, sentiment analysis, and joint sentiment/topic modeling of text data. In the third section, the theoretical foundations and the theory of the proposed model are explained. In this section, we use a well-known model as the basis of work to develop a new model and then describe its various parts and the relationships required in each part. In the fourth section, we explain the steps taken to evaluate the proposed model and then compare its performance in different experiments with other models. Also, we created two new datasets
 to evaluate the proposed model in the field of Information Retrieval. These two new datasets will be described in this section. In the final section, we present the conclusions of the paper and make some suggestions for improving and developing the proposed model.

\section{Related Works}
\label{sec2}
In this section, we review the literature related to topic modeling, estimation of probability distributions in input data, and joint sentiment/topic modeling based on the both neural networks and also Bayesian approaches.

The Restricted Boltzmann Machine (RBM) is an unsupervised two-layer neural network for the estimation of distribution of binary input data. This generative probabilistic model was first introduced in 1986 by Smolensky \cite{smolensky1986information} and was later developed by Hinton in 2002 \cite{hinton2002training}. Inspired by the RBM model, in 2011 Larochelle et al. introduced the Neural Autoregressive Distribution Estimation (NADE) \cite{larochelle2011neural}, which is an unsupervised generative probabilistic method for modeling the probability of discrete data. NADE eliminated the limitation of RBM in high dimensional joint probability estimations by the use of fully visible Bayes networks for probability calculations. The earliest neural network-based topic model is Replicated Softmax Model (RS) introduced by Hinton and Salakhutdinov in 2009 \cite{hinton2009replicated}, which is an extension of the RBM model used to detect the distribution of topics in text data \cite{hinton2009replicated}. In 2012, Larochelle and Lauly combined NADE and RS to develop an unsupervised neural network-based topic modeling method called the Document Neural Autoregressive Distribution Estimation (DocNADE) \cite{larochelle2012neural}.

In the category of Bayesian topic models, the well-known Latent Dirichlet Allocation (LDA) model introduced by Blei et al. in 2003 \cite{blei2003latent} has long served as the basis of all methods of this category. LDA is a generative probabilistic method in which text document is considered as a mixed distribution over topics, where each topic is characterized by a distribution over words.

All of the aforementioned models are capable of detecting the topics in text data. There is however another group of topic models that can detect the topics as well as the sentiments associated with each one. This group includes the Aspect-Sentiment Unification Model (ASUM) introduced by Jo and Oh in 2011 \cite{jo2011aspect} for detecting topics and sentiments in online reviews. ASUM is an extension of LDA and falls in the category of generative probabilistic graph models. In 2012, Lin et al. introduced the weakly supervised joint sentiment-topic (JST) detection model \cite{lin2012weakly}. The advantage of JST over its competitors is in its weakly supervised nature, which allows it to be easy adapted for other domains without noticeable decrease in performance.

\section{Proposed Model: Boosted RBM for Joint Sentiment/Topic Modeling}
\label{sec3}

\begin{figure}[!t]
	\centering
	\subfigure[RBM Model]{\label{fig1-1}\includegraphics[width=70mm]{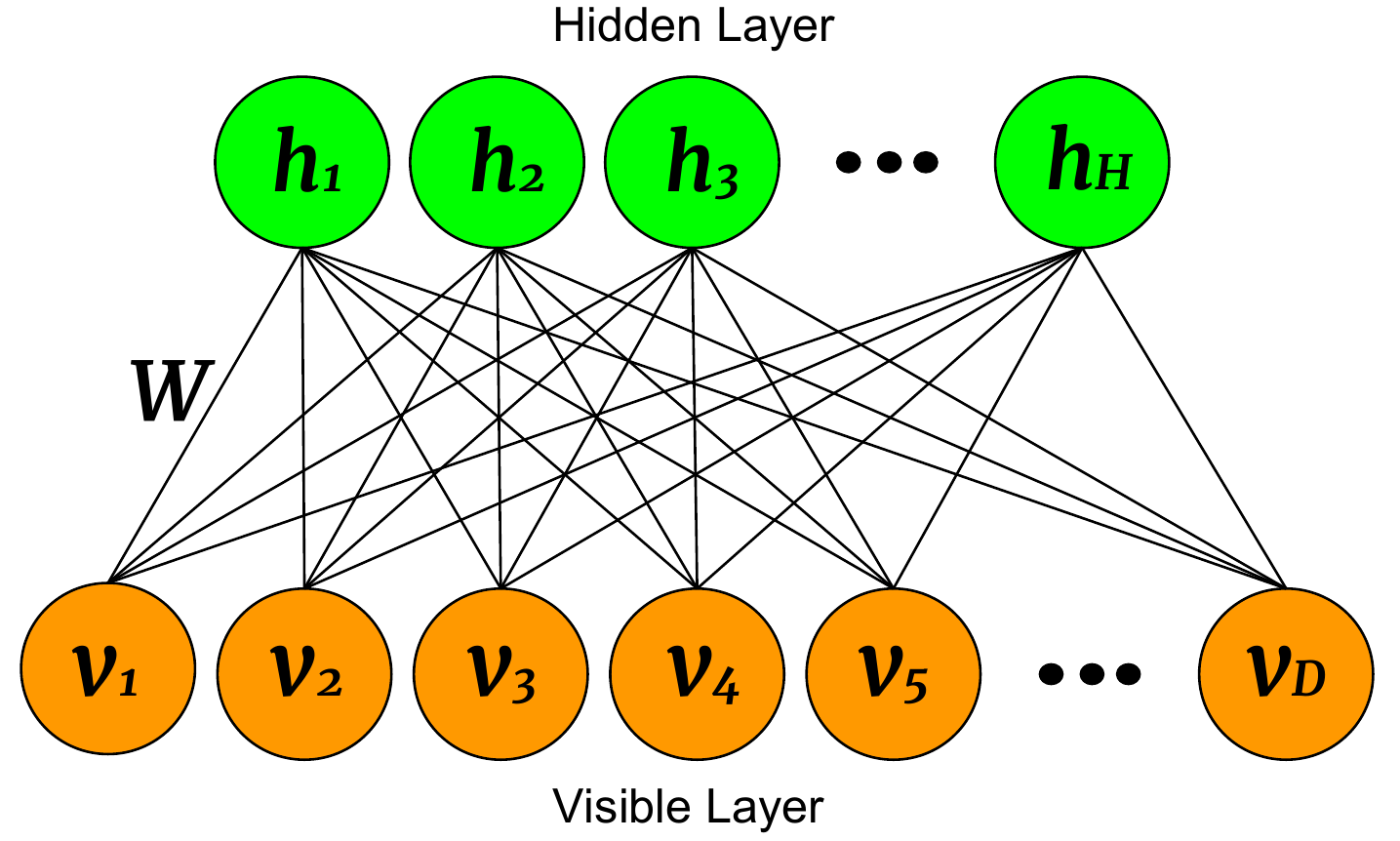}}
	\subfigure[RS Model]{\label{fig1-2}\includegraphics[width=70mm]{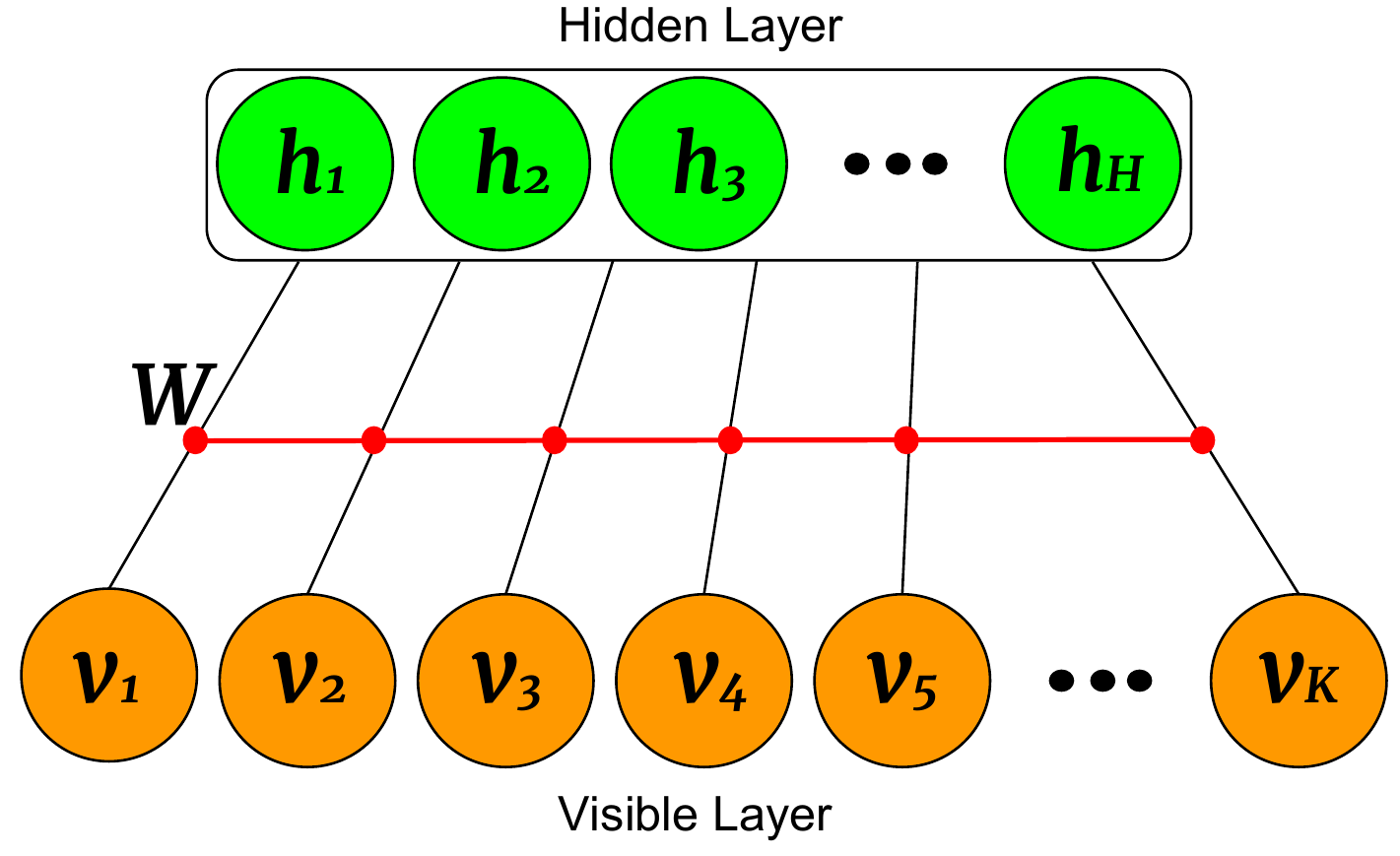}}
	\caption{RBM and RS Models}
	\label{fig1}
\end{figure}

The basis of the proposed model in this paper is the RBM \cite{hinton2002training} neural network shown in Fig. \ref{fig1-1}. In RBM, probability distributions of input data are obtained by minimization of an energy function defined as:
\begin{align}
\centering
\label{eq1}
E(\textbf{v},\textbf{h}) = -\sum_{i} \sum_{j} v_iW_{ij}h_j - \sum_i v_ia_i - \sum_j h_jb_j.
\end{align}

In Eq. \ref{eq1},  $\theta = \{W, \textbf{a}, \textbf{b}\}$ is the set of model parameters. $W_{D\times H}$ is the weight matrix for the connections between the input layer and the hidden layer, where $D$ is the size of the input vector, and $H$ is the size of the hidden layer. The parameter \textbf{a} is the bias vector of the input layer of size $D$ and the parameter \textbf{b} is the bias vector of the hidden layer of size $H$.

Now, assume that our aim is to utilize RBM to model the discrete data \textbf{v} where $ \textbf{v} \in \{1,...,K\}^D$. Here, $ K $ is the dictionary size, $ D $ is the document size, and
$\textbf{h} \in \{0,1\}^H$
is the hidden layer. We assume the matrix \textbf{V} of size $ K\times D $ as the visible binary matrix where $v_{ki}=1$ if visible unit $ i $ takes on $k^{th}$ value. the energy function for the state
$\{\textbf{V},\textbf{h}\}$
is defined as:
\begin{align}
\centering
\label{eq7}
E(\textbf{V},\textbf{h})=-\sum_{i=1}^{D}\sum_{j=1}^{H}\sum_{k=1}^{K}W_{kij}h_jv_{ki}-\sum_{i=1}^{D}\sum_{k=1}^{K}v_{ki}a_{ki} - \sum_{j=1}^{H}h_jb_j.
\end{align}
The conditional distributions are calculated in the form of softmax and logistic function:
\begin{align}
\centering
\label{eq8}
p(v_{ki}=1|\textbf{h})=\dfrac{exp(a_{ki}+\sum_{j=1}^{H}h_jW_{kij})}{\sum_{k=1}^{K}exp(a_{ki}+\sum_{j=1}^{H}h_jW_{kij})};\quad p(h_{j}=1|\textbf{V})=\sigma \left( b_j +\sum_{i=1}^{D} \sum_{k=1}^{K}v_{ki}W_{kij} \right)
\end{align}
\cite{hinton2009replicated}.

Now suppose, for each document, create an independent RBM with as many softmax units as there are words in the document. With the order of words ignored, all of the softmax units can share the weights that connect them to the hidden layer. Therefore, for a document consisting of $ D $ words, the energy function for the state
$ \{\textbf{V}, \textbf{h} \} $ is defined as:
 \begin{align}
 \centering
 \label{eq9}
 E(\textbf{v},\textbf{h})=-\sum_{j=1}^{H}\sum_{k=1}^{K}W_{jk}h_j\hat{v}_{k}-\sum_{k=1}^{K}v_{k}a_{k} - D\sum_{j=1}^{H}h_jb_j
 \end{align}
where $\hat{v}^k = \sum_{i=1}^{D}v_{ik}$. Equation \ref{eq9} is in fact the RS model proposed by Hinton and Salakhutdinov \cite{hinton2009replicated} and shown in Fig. \ref{fig1-2}.

\begin{figure}[!t]
 	\centering
 	\includegraphics[scale=0.5]{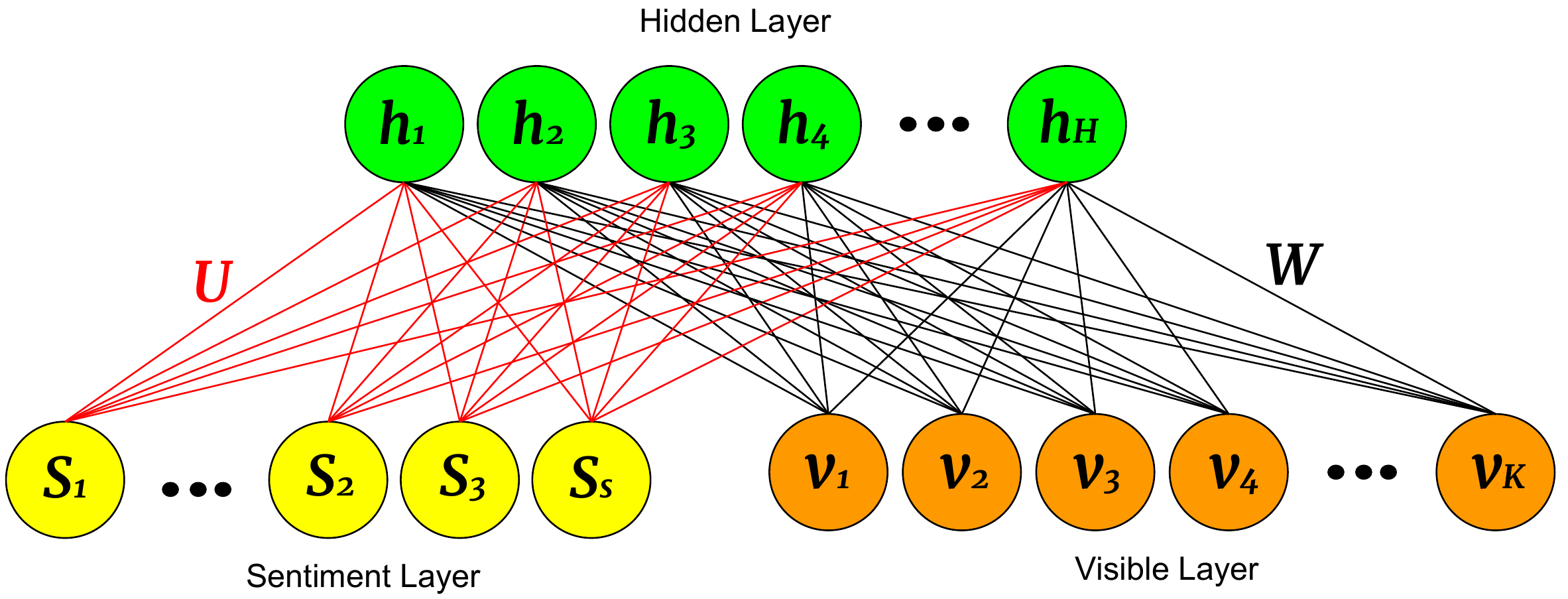}
 	\caption{Generative Probabilistic Proposed sentiment/topic Model}
 	\label{fig3}
\end{figure}

The above formulations constitute the basis of the proposed structure. The model of this paper is developed by extending the above formulations as described below. The proposed model is a RBM-based generative probabilistic model for sentiment/topic modeling of text data. As shown in Fig.  \ref{fig3}, like RBM, this method has a two-layer structure, but with a vector corresponding to the document label or the number of existing classes added to the visible layer of the structure. The input vector of this structure in the visible part is a constant-length vector of the same size as the dictionary, where the number of word repetitions is specified.

As shown in Fig. \ref{fig3}, for each text document, the proposed model receives a binary vector representing the sentiment of document as input. The existing distributions over words in each topic and their associated sentiments are extracted in the hidden layer. In the presence of the additional layer and its associated parameters, the energy calculation equation is turned into:
 \begin{align}
 \label{eq12}
 E(\textbf{V},\textbf{s},\textbf{h})=-\sum_{j=1}^{H}\sum_{k=1}^{K}W_{kj}h_j\hat{v}_{k}&-\sum_{j=1}^{H}\sum_{l=1}^{S}U_{lj}h_js_l\\\nonumber
 &-\sum_{k=1}^{K}v_{k}a_{k} -\sum_{l=1}^{S}s_lc_l- D\sum_{j=1}^{H}h_jb_j
 \end{align}
In Eq. \ref{eq12}, $\theta=\{W, U, \textbf{a}, \textbf{b}, \textbf{c} \}$
is the set of model parameters where $W_{K\times H}$ is the weight matrix for the connection between the visible layer and the hidden layer,  $U_{S\times H}$  is the weight matrix for the connection between the sentiment layer and the hidden layer, and $\textbf{a}$, $\textbf{b}$ and $\textbf{c}$ are the bias vectors of the visible, hidden, and sentiment layers, respectively. $ K $ and $ H $ are the sizes of the dictionary and the hidden layer, and $ S $ is defined as the number of existing sentiments or the size of the sentiment vector. The probability that the model assigns to each document and its associated sentiment layer is calculated as follows:
\begin{align}
\centering
\label{eq13}
p(\textbf{v},\textbf{s},\textbf{h}) = \dfrac{1}{Z} e^{-E(\textbf{v},\textbf{s},\textbf{h})} \Rightarrow
p(\textbf{v},\textbf{s}) = \dfrac{1}{Z} \sum_{h}  e^{-E(\textbf{v},\textbf{s},\textbf{h})} \ \ ,
Z = \sum_{\textbf{v}}\sum_{\textbf{s}}\sum_{\textbf{h}} e^{-E(\textbf{v},\textbf{s},\textbf{h})}.
\end{align}
In Eq. \ref{eq13}, $Z(\theta)$ is the partition function and ensures that the value obtained for configuration  $(\textbf{v}, \textbf{s}, \textbf{h})$ in Eq. \ref{eq13} is an integer between 0 and 1.

In the proposed model, the values of visible, sentiment and hidden layers are calculated as follows:
\begin{align}
\centering
\label{eq15}
p(v_{i}=w|\textbf{h})=\dfrac{exp(a_{w}+\sum_{j=1}^{H}W_{wj}h_j)}{\sum_{k=1}^{K}exp(a_{w}+\sum_{j=1}^{H}W_{wj}h_j)}; \quad p(s_{l}=1|\textbf{h})=\dfrac{exp(c_{l}+\sum_{j=1}^{H}U_{lj}h_j)}{\sum_{l=1}^{S}exp(c_{l}+\sum_{j=1}^{H}U_{lj}h_j)}
\end{align}
\begin{align}
\centering
\label{eq17}
p(h_{j}=1|\textbf{v},\textbf{s})=\sigma \left( Db_j + \sum_{k=1}^{K}W_{kj}\hat{v}_k + \sum_{l=1}^{S}U_{lj}s_l \right)
\end{align}

The value of the hidden layer depends on the values of both visible and sentiment layers. Therefore, in Eq.  \ref{eq17}, the value of the hidden layer is obtained by sampling a conditional distribution depending on the values of both visible and sentiment layers. The reason for the use of softmax function for the visible and sentiment layers in Eq. \ref{eq15} is that once calculated the values of these layers conditioned to the hidden layer, need to be sampled related to these values. The use of softmax function ensures that the values calculated for these two vectors are a polynomial probability distribution that can be easily sampled.

\subsection{Training of the Proposed Model}
\label{sec3-1}
The CD algorithm
\cite{hinton2002training}\cite{woodfordnotes}\cite{hinton2010practical}\cite{carreira2005contrastive}
is used to train the proposed model and update the network parameters, including the weight matrices for the connections between the visible and hidden layers and between the sentiment and hidden layers, as well as the biases of all three layers. The model parameters are updated with the following equation:
\begin{align}
\centering
\label{chap4-eq28}
\triangle\theta = \alpha \left( E_{P_{data}}[\theta]-E_{P_{model}}[\theta]\right) \Rightarrow \theta_{t+1} = \theta_t + \triangle\theta.
\end{align}
In Eq. \ref{chap4-eq28}, $E_{p_{data}}[.]$ is the expected value of the model parameters according to data distribution and $E_{p_{model}}[.]$ is the expected value of the model parameters according to the distribution obtained by the model.

\section{Experiments}
\label{sec4}
In this section, we explain the procedure of model testing and evaluation, and report and analyze the results obtained from different tests. The purpose of these tests is to observe the effect of adding a sentiment layer on topic modeling, sentiment tagging, classification, and information retrieval. In Sections \ref{sec4-5} and \ref{sec4-6}, the proposed approach is compared with the RS model.

\subsection{Description of Datasets}
\label{sec4-1}
Our tests and evaluations were performed by the use of several standard datasets of the field of topic modeling and sentiment analysis. A brief description of these databases is provided in the following.

The 20-Newsgroups (20NG) \cite{20ng} dataset is a well-known dataset in the field of topic modeling. This dataset consists of 18,786 text documents collected from Usenet newsgroup repositories. This document collection is divided into 20 newsgroups, each related to one specific topic. Of the 18,786 documents in this dataset, we use 11,284 documents for the training and use 7,502 documents for testing the trained model. The 2000 most frequently repeated words in this dataset are used to compile the dictionary.

The movie review (MR) \cite{mr} dataset compiled by Pang et al. \cite{lin2012weakly}\cite{pang2002thumbs} is another standard dataset for performance evaluation in the field of topic modeling. Our tests are conducted using the second version of this dataset, which includes 1,000 positive and 1,000 negative movie reviews collected from the Internet Movie Database (IMDB) website. The average length of each review in this dataset is 30 sentences.

The third dataset is the multi-domain sentiment (MDS) dataset introduced by Blitzer et al. in 2007 \cite{blitzer2007biographies}, which consists of collected reviews about four types of Amazon products: Books, DVDs, electronics, and kitchen appliances. The MDS dataset contains 1,000 positive and 1,000 negative reviews for each of the above mentioned product types.

\subsection{Preparation of Datasets}
\label{sec4-2}
After preprocessing the texts of the MR dataset (removing stop words, stemming and lemmatizing), each document was converted into a sequence of words. In addition to the dictionary compiled from preprocessed data (with a size of 24,916 words), we also used 2000-word and 10000-word dictionaries belonging to the 20NG and Reuters Corpus Volume I (RCV1) \cite{rcv1} datasets, respectively. We called these three states of Movie Review dataset MR1, MR2 and MR3. The statistics obtained from the described procedures are presented in Table \ref{tb2}. In the next step, we partitioned the database into two subsets, one for training and another for testing. Each of these subsets consisted of 1000 documents, 500 with positive tag and 500 with negative tag.
\begin{table}[!t]
	\footnotesize
	\centering
		\begin{tabular}{|c|c|c|c|c|c|}
			\hline
			Dataset Name & Dictionary Size & Num of Train & Num of Test & Avg Docs Length & Std Deviation \\
			\hline
			MR1 & 2000 & 1000 & 1000 & 90.18 & 40.23 \\
			\hline
			MR2 & 10000 &1000 & 1000 & 186.35 & 81.33 \\
			\hline
			MR3 & 24916 &1000 & 1000 & 299.75 & 126.51 \\
			\hline
		\end{tabular}
	\caption{Movie Review Dataset Statistical Information}
	\label{tb2}
\end{table}

\subsection{Sentiment Lexicon}
\label{sec4-3}
The sentiment lexicon is a pre-made general dictionary where for each word there are three sentiment tags, positive, negative, and neutral, each assigned with a weight between 0 and 1 so that the sum of all weights is 1. In this paper, we use a sentiment dictionary called MPQA \cite{mpqa}. This sentiment dictionary contains 4053 words, each with a 3-element vector, where the first element represents the neutrality weight, the second element represents the positivity weight, and the third element represents the negativity weight of the corresponding word. Overall, this dictionary contains 1511 positive words and 2542 negative words.

\subsection{Details of Training}
\label{sec4-4}
We used the MR database as the input data to train the model in three different modes (MR1, MR2 and MR3) before testing its performance. The results of the conducted tests are presented later in the paper. The training on the all three states was performed using the first order CD algorithm. In all three training modes, the model was trained for 1000 iterations on the entire training subset with batch size of 1. The other parameter involving the training is the number of units in the hidden layers (h), which equals the number of topics. In all training modes, we trained the proposed method and the RS model for $h=\{5,10,15,20,25,30,35,40,45,50,60,70,80,90 \}$. For all training modes, we used learning coefficient $ \alpha = 0.001$. The parameters $ W $ and $ U $, which represent the weights of connections between the visible and hidden layers and between the sentiment and hidden layers, and the parameters $ \textbf{a}  $ and $ \textbf{c} $, which are the biases of the visible and sentiment layers, were initialized with random numbers from a Gaussian distribution with a mean of 0 and variance 1. The bias of the hidd	en layer $ \textbf{b} $ was initialized at zero.

\subsection{Document Modeling and Evaluation as a Generative Model}
\label{sec4-5}
In this section, we present the results of performance evaluation of the proposed method as a generative probabilistic model in comparison with the RS model. As mentioned, these evaluations were performed after training with three dictionaries and using the documents in the test set. Through the analysis of the results, we show that the proposed method outperforms the RS model in probability estimation for the unobserved documents.

We use a criterion called perplexity to evaluate the calculated probability for the documents. Perplexity is a commonly used criterion for comparison of different probabilistic models in the NLP field. Perplexity has been defined as:
\begin{align}
\centering
\label{eq18}
Perplexity = exp\left( - \dfrac{\sum_{n=1}^{N}\log p(\textbf{v}_n)}{\sum_{n=1}^{N}D_n} \right).
\end{align}
According to the Eq. \ref{eq18}, perplexity equals the inverse of the mean per word likelihood obtained for each document on log-scale. In the modeling with an appropriate probabilistic model, perplexity should be monotonically decreasing. Overall, the lower is the model perplexity on the dataset, the better is the model quality.
\begin{figure}[!t]
	\centering
	\subfigure[MR1]{\label{fig4-1}\includegraphics[width=55mm]{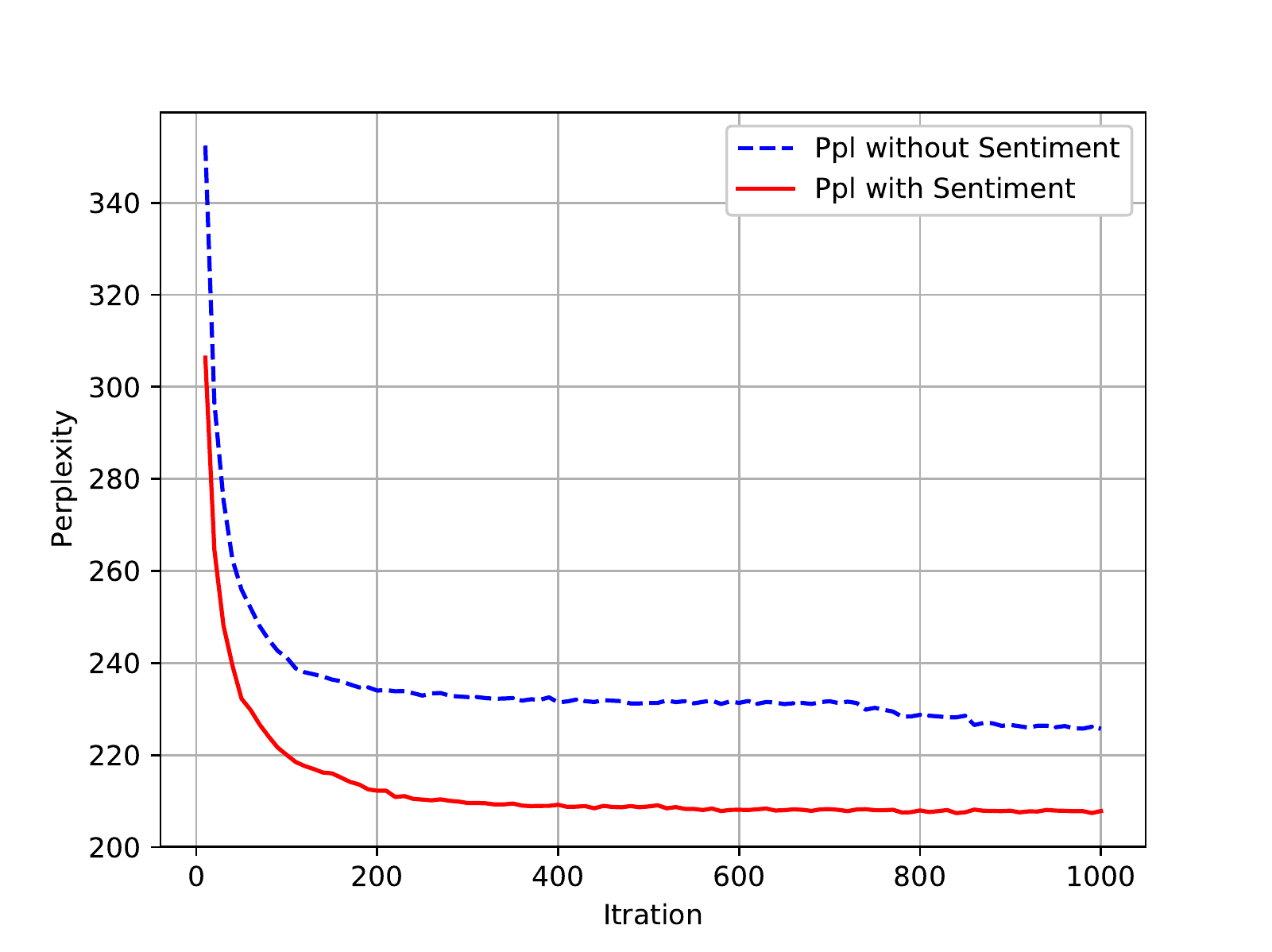}}
	\subfigure[MR2]{\label{fig4-2}\includegraphics[width=55mm]{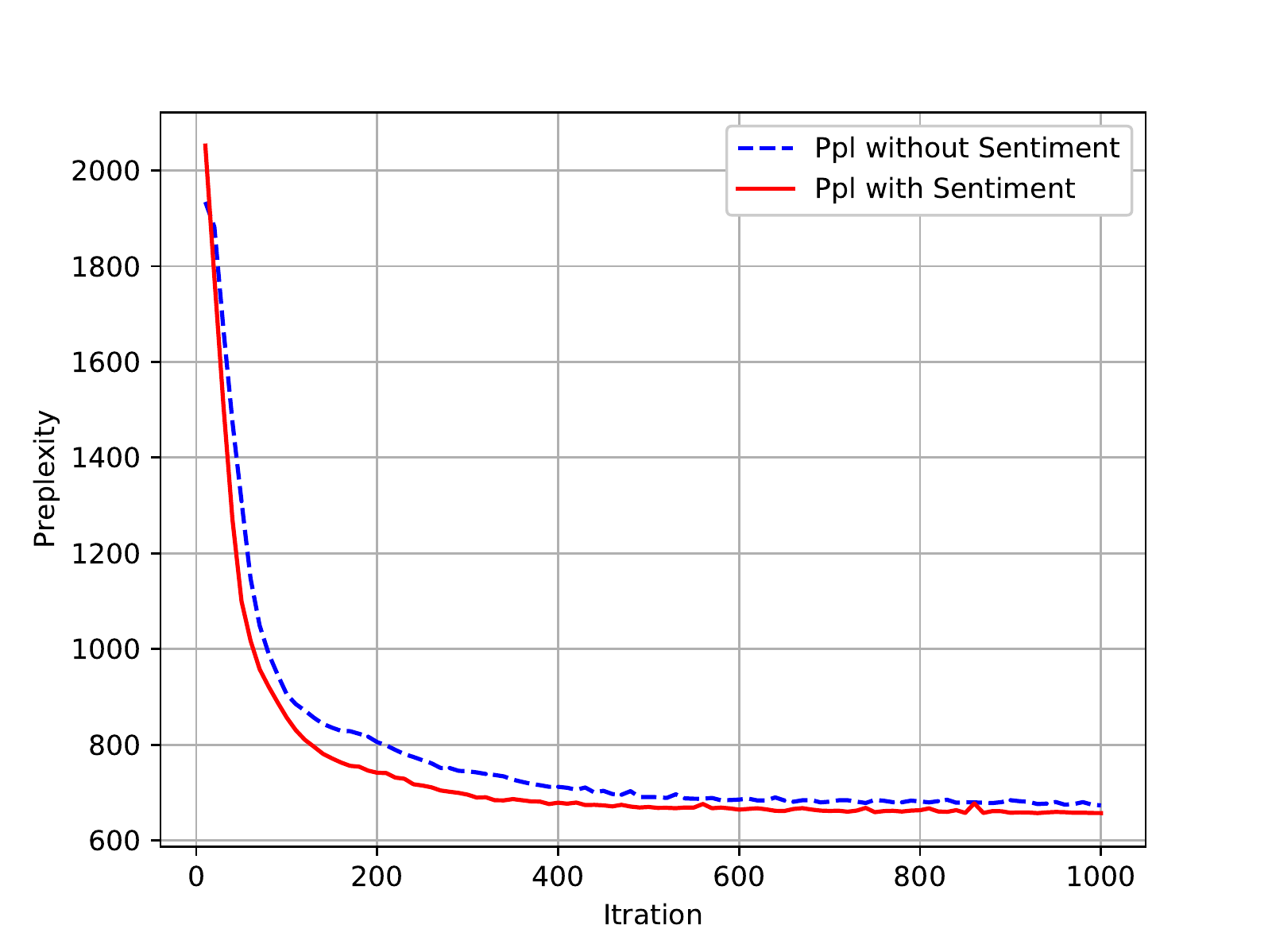}}
	\subfigure[MR3]{\label{fig4-3}\includegraphics[width=55mm]{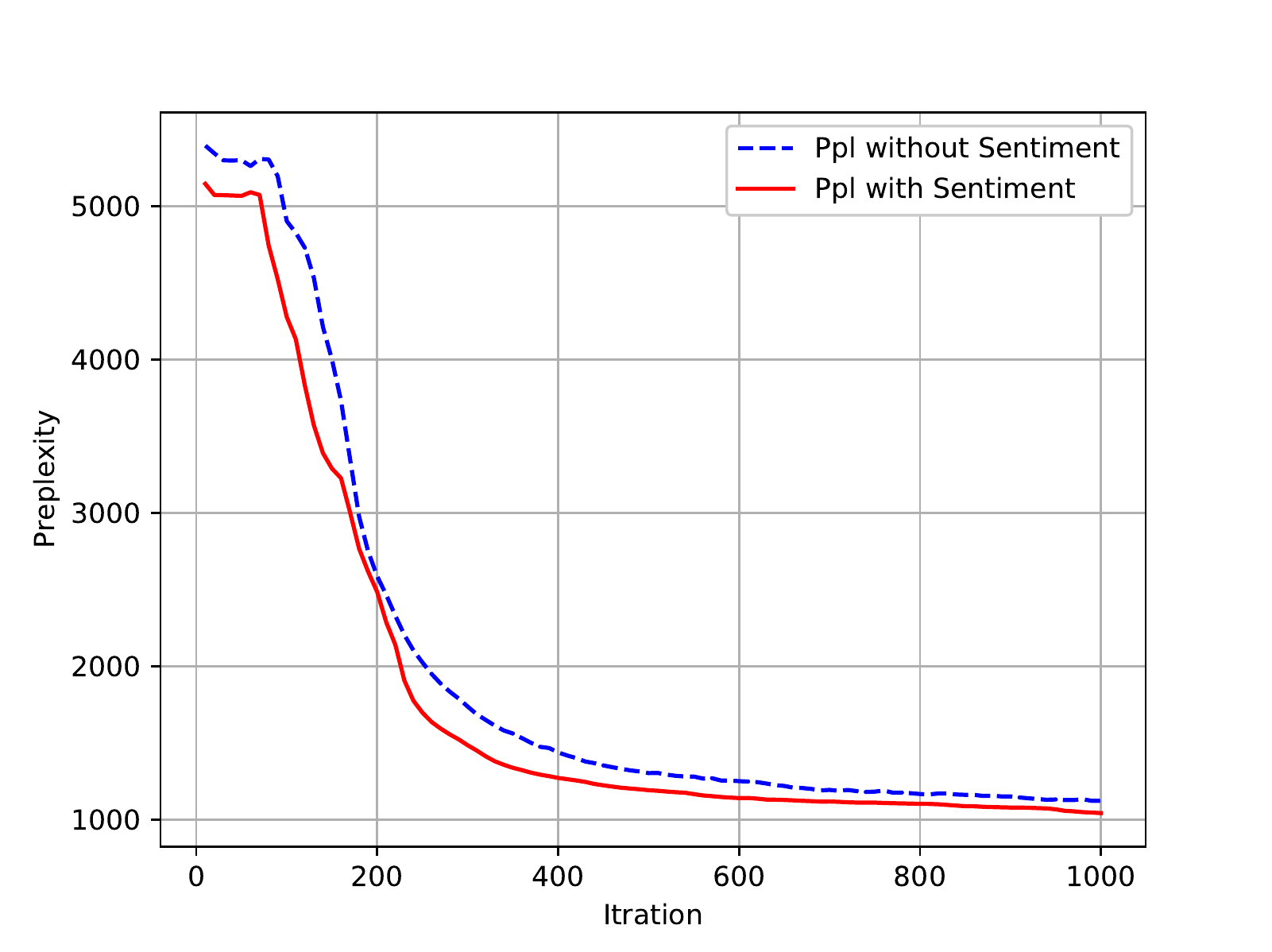}}
	\caption{Evaluation of perplexity variation during the training on MR dataset}
	\label{fig4}
\end{figure}

Figure \ref{fig4} shows the variation of perplexity of the proposed model and the RS model during the training with the MR dataset. As can be seen, in all three charts, the proposed joint sentiment/topic model has a greater perplexity decline than the RS topic model. Also, in all three charts, perplexity decline is sharper at the beginning of the training than at the final stages. This is so that from the 200th iteration onward, there is no significant change in perplexity. Careful examination of Fig. \ref{fig4} reveals that adding the sentiment layer to construct a generative probabilistic model, as we did in this paper, leads to greater perplexity reduction in the training phase and therefore to development of a better probabilistic method for document modeling.

The calculated perplexity values presented in Table \ref{tb3} are also the proof of higher performance of the proposed generative approach in the modeling process. The perplexity values shown in Table \ref{tb3} are for the test set of the MR dataset and 2000-word, 10000-word, and 24916-word dictionaries. As shown in Table \ref{tb3}, the perplexity values obtained for the proposed model are lower than those obtained for the RS model. Thus, as stated earlier, the use of an additional layer dedicated to sentiment leads to development of a probabilistic document modeling method capable of outperforming the RS method.
\begin{table}[!b]
	\footnotesize
	\centering
		\begin{tabular}{|c|c|c|}
			\hline
			TestSet Type  & Ppl without Sentiment & Ppl with Sentiment \\
			\hline
			MR1 & 423.89 & \textbf{406.74} \\
			\hline
			MR2  & 2028.69 & \textbf{1871.57} \\
			\hline
			MR3  & 5842.39 & \textbf{5824.97}\\
			\hline
		\end{tabular}
	\caption{Perplexity Estimation on Movie Review Dataset Using Proposed Model }
	\label{tb3}
\end{table}

\subsection{Information Retrieval}
\label{sec4-6}

Since the proposed approach is a generative method for simultaneous modeling of topics and sentiments, the first requirement for the evaluation of this model in the data retrieval context is to use a dataset with both sentiment and topic labels for every document. Given the absence of such dataset, we created two datasets with both sentiment and topic tags for the testing purpose.

The first sentiment/topic database was created by assigning sentiment tags to the 20NG dataset. To do so, for each document we counted the number of words with known sentiment polarity using the MPQA sentiment dictionary. Then, the documents for which the number of positive words was greater than negative were given a positive tag and vice versa. We called this dataset S-20NG.

The second database created for the evaluation of the proposed method in the information retrieval context was created by compilation of the MR and MDS datasets introduced in Section \ref{sec4-1}. All of these 5 datasets (MDS alone consists of 4 different parts, each containing 2000 documents) only have sentiment labels. But each of these documents can be considered to represent a specific topic. Thus, these datasets were combined together to create a new larger dataset called MRMDS,
which consists of 10000 documents, 5000 with positive tags and 5000 with negative tags, and five topics including: movie, book, DVD, electronic, and kitchen appliances. After the preprocessing phase, each of these documents was converted to the lib-svm file using the 2000-word dictionary of the 20NG dataset. Of the 10000 document obtained by combining these 5 datasets, 7500 documents with even sentiment label distribution (3750 documents labels with positive and 3750 documents with negative labels) and even topic distribution (1500 documents -consisting of 750 positive and 750 negative documents- per topic,) were assigned to the training set. The remaining 2500 documents (500 documents -consisting of 250 positive and 250 negative documents- per topic) were assigned to the testing set.

\begin{figure}[!t]
	\centering
	\subfigure[S-20NG Dataset and Hidden Layer with Size10]{\label{fig5-1}\includegraphics[width=80mm]{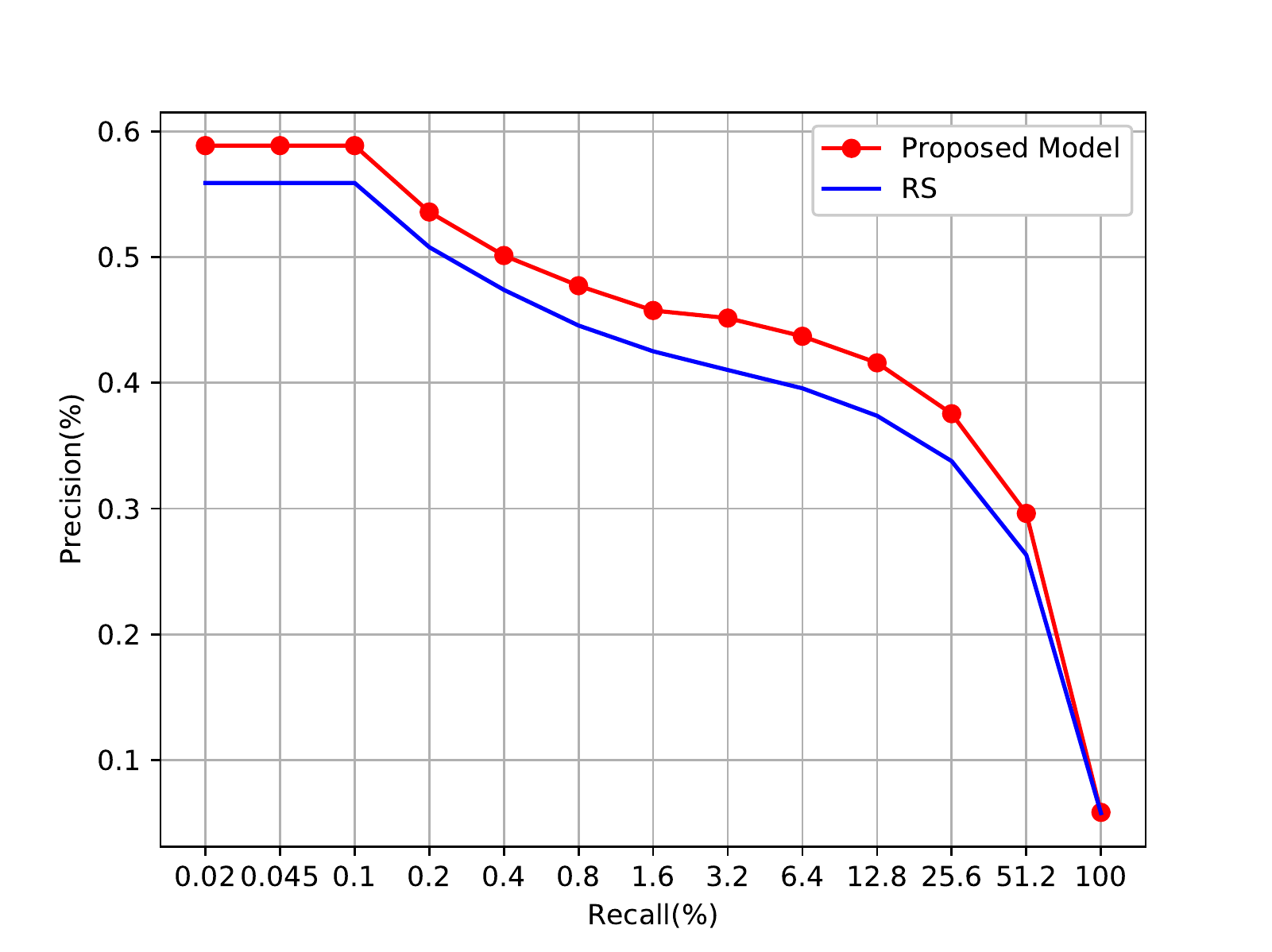}}
	\subfigure[MRMDS Dataset and Hidden Layer with Size 10]{\label{fig5-2}\includegraphics[width=80mm]{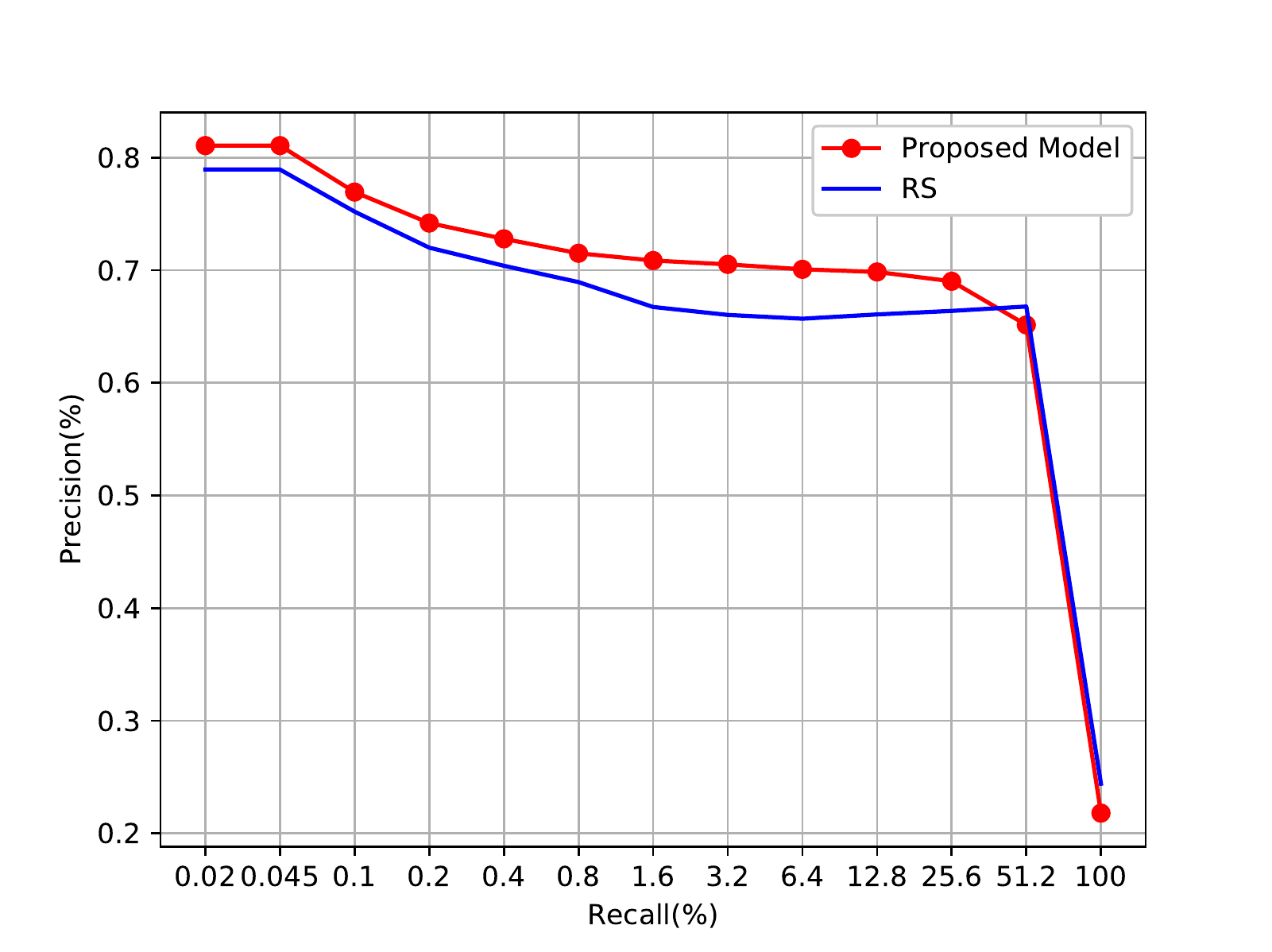}}
	\subfigure[S-20NG Dataset and Hidden Layer with Siz 50]{\label{fig5-3}\includegraphics[width=80mm]{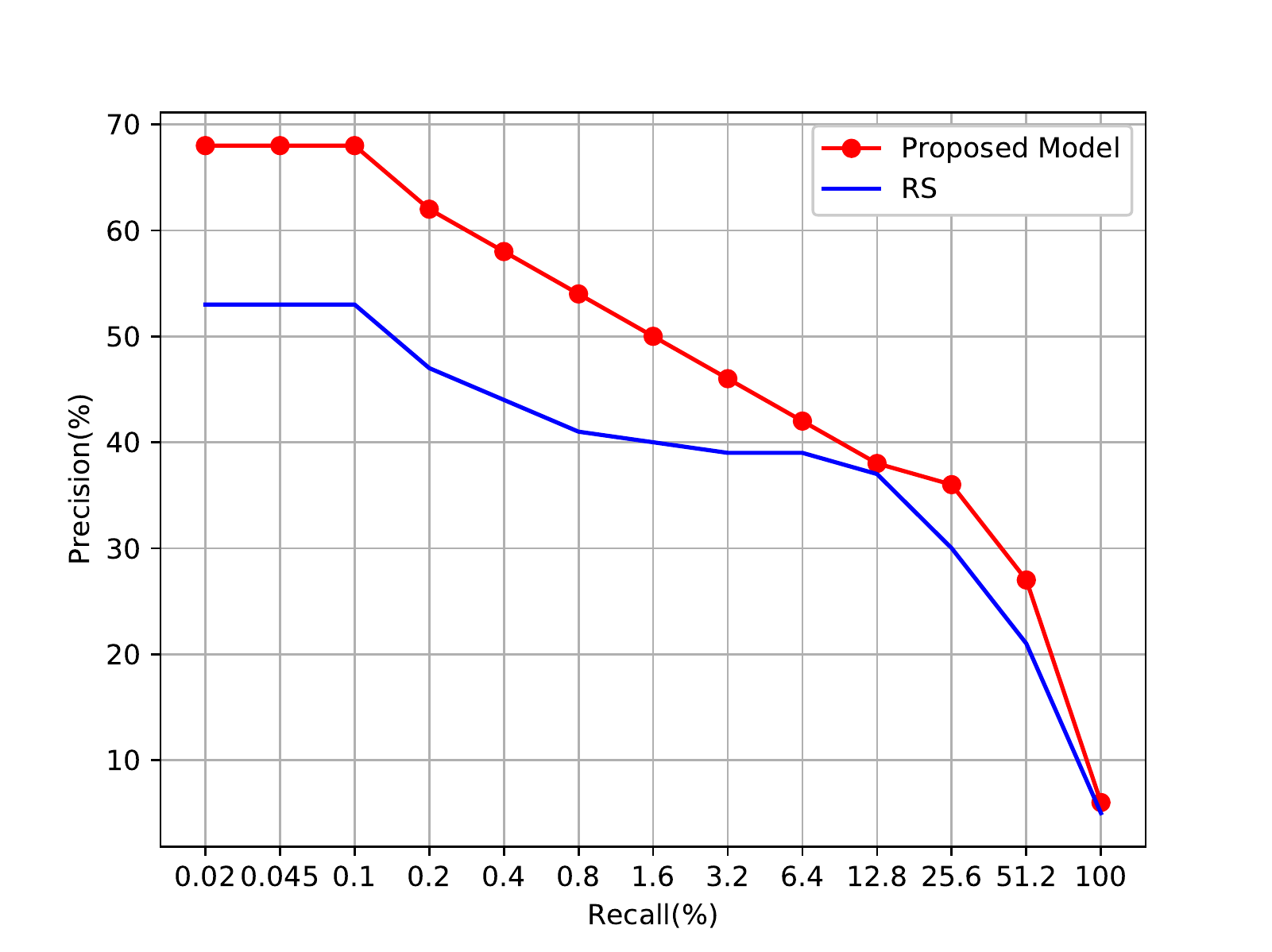}}
	\subfigure[MRMDS Dataset and Hidden Layer with Size 50]{\label{fig5-4}\includegraphics[width=80mm]{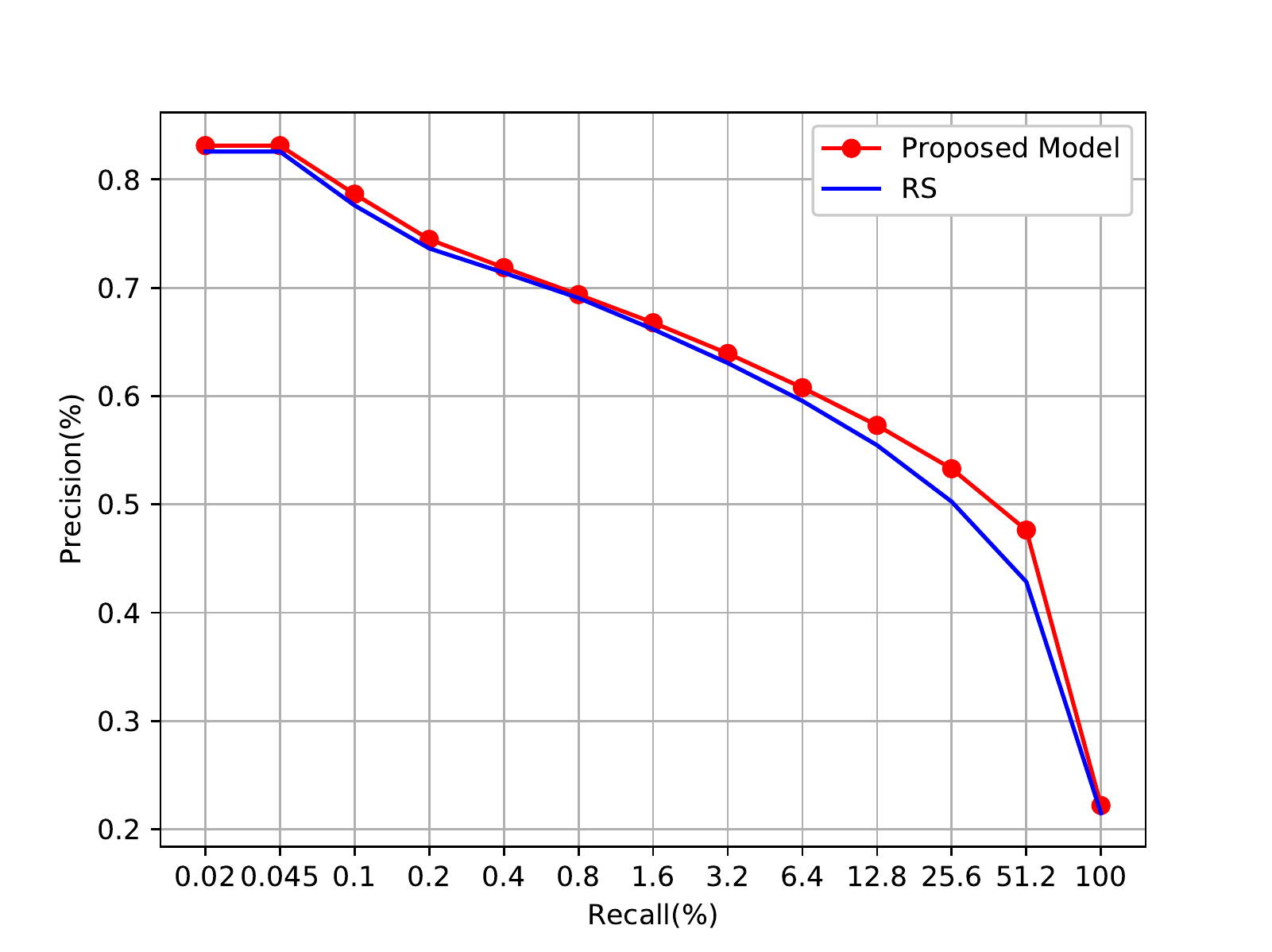}}
	\caption{Information Retrieval on S-20NG and MRMDS Datasets by Using Proposed Model and RS Model}
	\label{fig5}
\end{figure}

The purpose of the evaluation is to assess the effects of considering sentiment on the information retrieval with the proposed model. For this evaluation, we used the precision-recall plot. This plot is a well-known criterion for the evaluation and comparison of information retrieval methods. To obtain this plot, the precision and recall values achieved by each model must be plotted against each other.

Figures \ref{fig5} shows the results obtained by the evaluation of data retrieval performance of the proposed approach and the RS model. As can be seen in both charts of Fig. \ref{fig5}, and especially in Chart \ref{fig5-1}, the proposed method has a better data retrieval performance than the RS model. The precision and recall values plotted in Fig. \ref{fig5} were obtained as described in the following. First, we trained the proposed model only with sentiment labels (without topic labels) for 500 iterations using the hidden layers of size 10 and 50 units. For the RS model, training was performed without any labels for 500 iterations using the hidden layers of size 10 and 50 units. Then, for each document in each testing set, we calculated the cosine similarity of document with all documents of the training set to obtain precision and recall rates. Finally, the precision values obtained for the entire testing set was averaged and the charts of Fig. \ref{fig5} were plotted.

\subsection{Topic Visualization}
\label{sec4-7}
This section presents the results obtained by the use of MPQA sentiment dictionary to evaluate the precision of topic models in terms of sentiment label assignment. This evaluation is inspired by the test conducted elsewhere on well-known topic models such as DocNADE \cite{larochelle2012neural} and LDA \cite{blei2003latent}.

Given the structure of the proposed approach (described in Section \ref{sec3}), we know that each hidden layer unit is connected to all units both in the visible layer and in the sentiment layer. Each unit in the sentiment layer is equivalent to a sentiment tag, and each unit in the visible layer corresponds to a word. In the topic modeling of text documents, each topic is defined as a polynomial probability distribution on all dictionary words, so we know that each unit of the hidden layer is connected, with a specific weight, to all dictionary words in the visible layer. For each word, this weight represents the significance of that word in that topic.

\begin{table}[!t]
	\footnotesize
	\centering
		\begin{tabular}{|c|c|c|c|}
			\hline     & Total Number & Numb of Positive Words & Num of Negative Words \\ \hline
			NG(2000)  &     155      &          100           &          55           \\ \hline
			RCV(10000) &     950      &          447           &          503          \\ \hline
			MR(24916)  &     3114     &          1242          &         1872          \\ \hline
		\end{tabular}
	\caption{Number of words shared between each of the three dictionaries and the MPQA sentiment lexicon}
	\label{tb4}
\end{table}

For this evaluation, we first calculated the total number of words shared between each of the three dictionaries and the MPQA sentiment lexicon. Table \ref{tb4} shows the results obtained from this operation. Then, we followed the below procedure for each of the modeling modes and topic numbers:
\begin{enumerate}
\item Calculating the total weights of positive words and negative words for each topic by the use of the sentiment dictionary and the matrix weight for the connection between the visible and hidden layers.
\item Calculating, for each topic, the difference between the two values calculated in step 1 and sorting the answers in descending order.
\item Assigning positive tags to the top five topics of the ordered list (most positive topics); and assigning negative tags to the bottom five topics of this list (most negative topics).
\item Comparing the tags assigned to each topic with the corresponding topic weights in the connection of the sentiment layer to calculate the precision.
\end{enumerate}

\begin{figure}[!t]
	\centering
	\includegraphics[width=70mm]{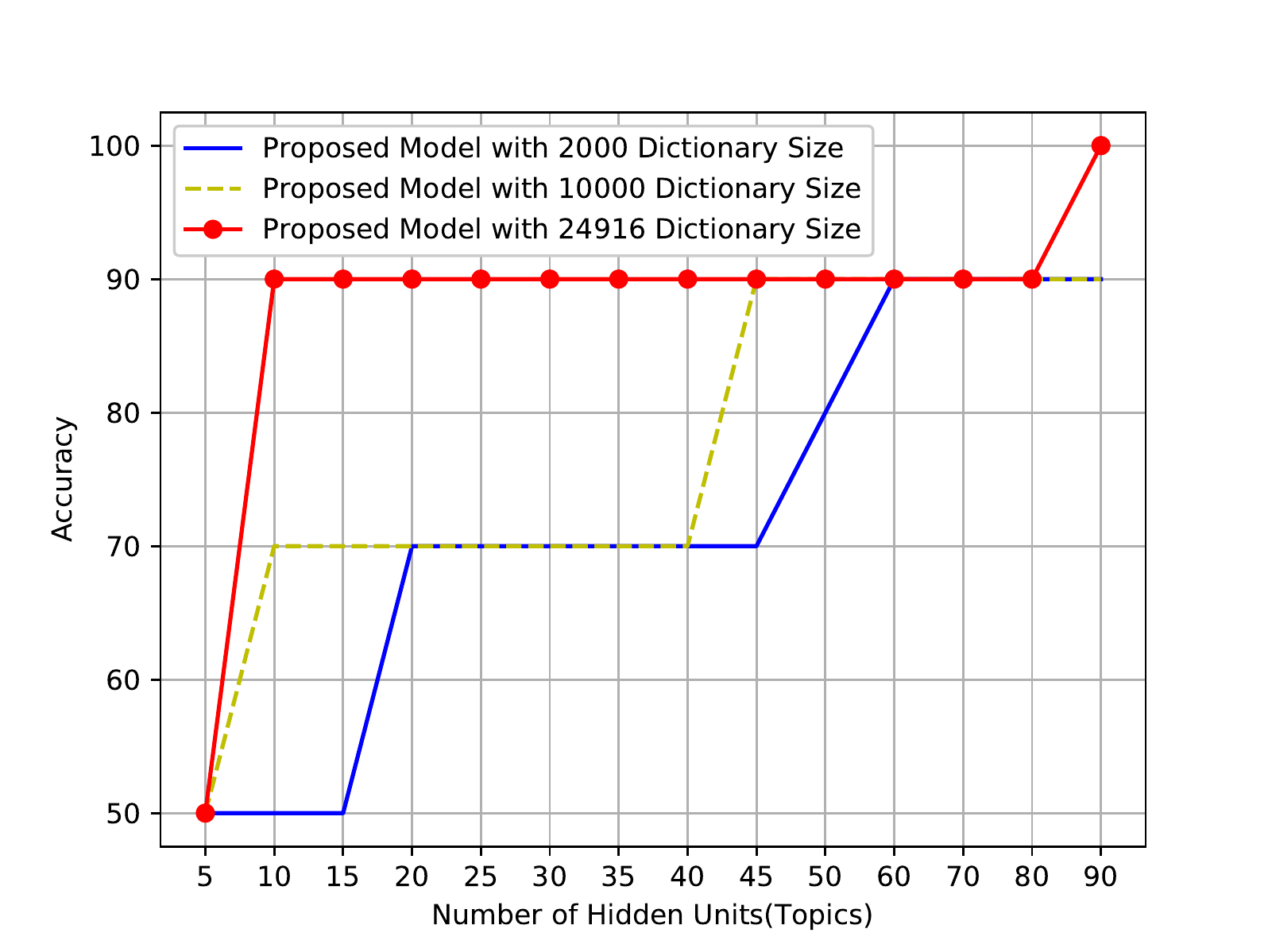}
	\caption{Precision Evaluation on Sentiment Assignment to Topics }
	\label{fig6}
\end{figure}

The idea behind the comparison made in step 4 (comparison of the tag assigned to each topic with the corresponding weight in the sentiment layer) is that for a topic assigned with a positive tag in step 3, the weight corresponding to the positive sentiment tag for that topic in the sentiment layer should be greater than the negative weight for the same topic and vice versa. Figure (5) shows the results of this evaluation. This figure indicates that as the dictionary size increases, so does the model precision in the assignment of sentiment tags to the topics.
A comparison of the values presented for different dictionaries in Table \ref{tb4} with Fig. \ref{fig6} reveals the cause of the relationship between the model precision and the dictionary size. As can be seen, as the dictionary size increases, so does the number of words shared between dictionary and the sentiment dictionary, and this leads to greater differentiation of the positive and negative topics in the training process, which result in improved model precision in the training and in assigning sentiment label to the topics.

\subsection{Sentiment Classification}
\label{sec4-8}
This section presents the results of the sentiment classification performed for the MR dataset using the proposed approach.  We use a basic word count-based method to evaluate the sentiment classification precision of the proposed method in different modes. We also use the sentiment classification results obtained from the support vector machine (SVM) and two neural networks, one initialized with random values and another initialized with the values given by the proposed method after training, to evaluate the parameters learned by the model.

Both neural networks used for comparison are of MLP type and utilize cross-entropy error function. In both networks, the number of neurons in the first and second layers is equal to the number of topics and sentiments respectively. Also, the first and second layers of both networks operate based on the tanh activation function and the softmax function, respectively.

To calculate the precision of the basic model, we counted the words with a specific sentiment polarity in each document of the test set. In other words, for each document, we calculated the number of positive words and negative words using the MPQA sentiment lexicon. After listing the number of positive and negative words for each document, we assigned a positive label to any document for which positive word count was greater than negative, and assigned a negative label to the documents with the opposite property.

For sentiment classification using the proposed approach, we first used the following equation:
\begin{align}
\centering
\label{eq11}
p(h_{j}=1|\textbf{V})=\sigma \left( Db_j + \sum_{k=1}^{K}W_{kj}\hat{v}_k \right)
\end{align}
to obtain for each text document, the probability value of each hidden unit. The next step was to calculate the sentiment layer corresponding to the current document using Eq. \ref{eq15}. Since the value of this layer is given by a softmax function, it is in the form of a probability distribution where entries add up to 1. Then, for each document, we checked the values obtained for the sentiment layer, and assigned the document with the sentiment label corresponding to the greatest value observed in that layer.

\begin{figure}[!b]
	\centering
	\subfigure[Dictionary Size 2000]{\label{fig7-1}\includegraphics[width=70mm]{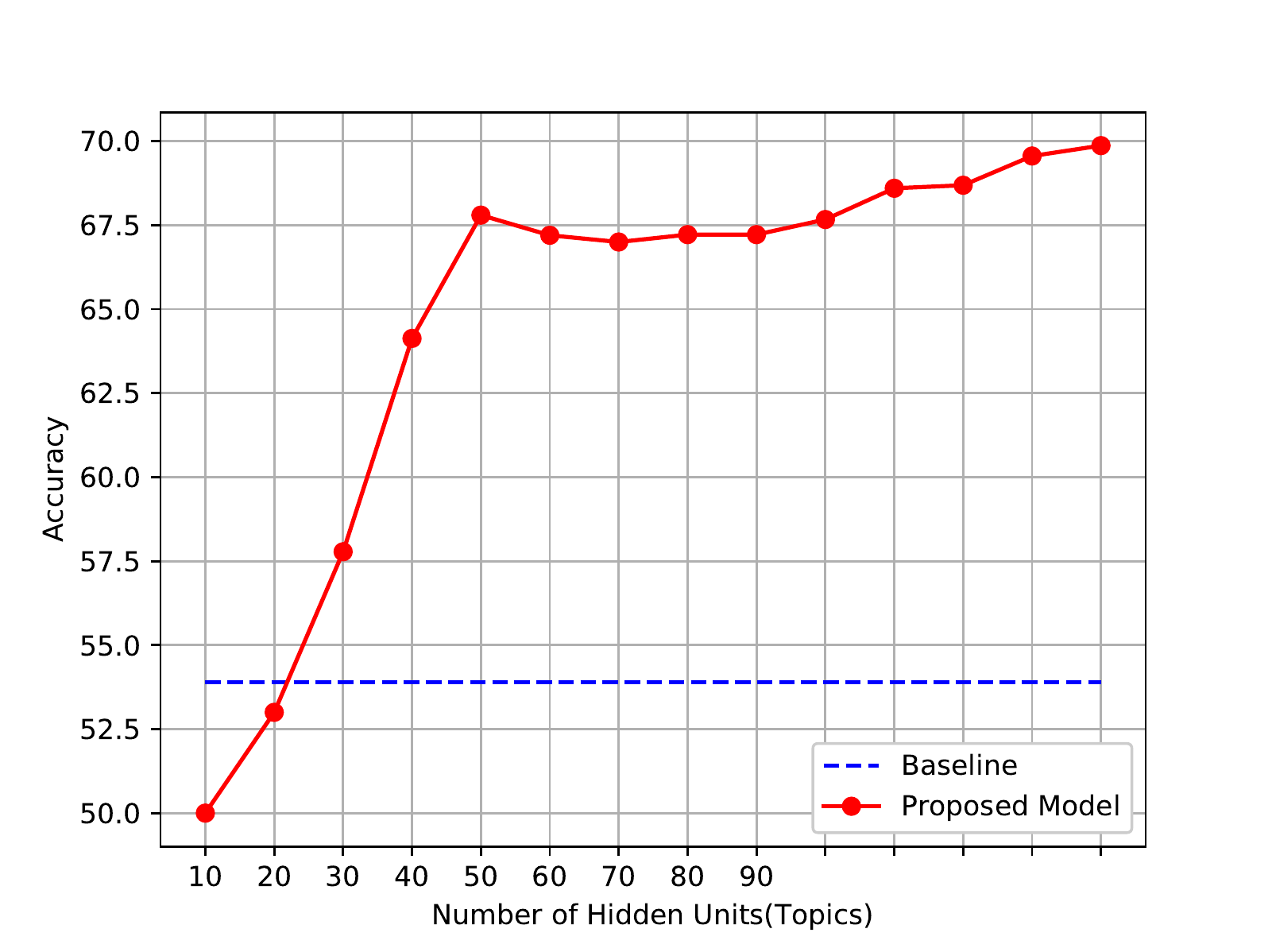}}
	\subfigure[Dictionary Size 10000]{\label{fig7-2}\includegraphics[width=70mm]{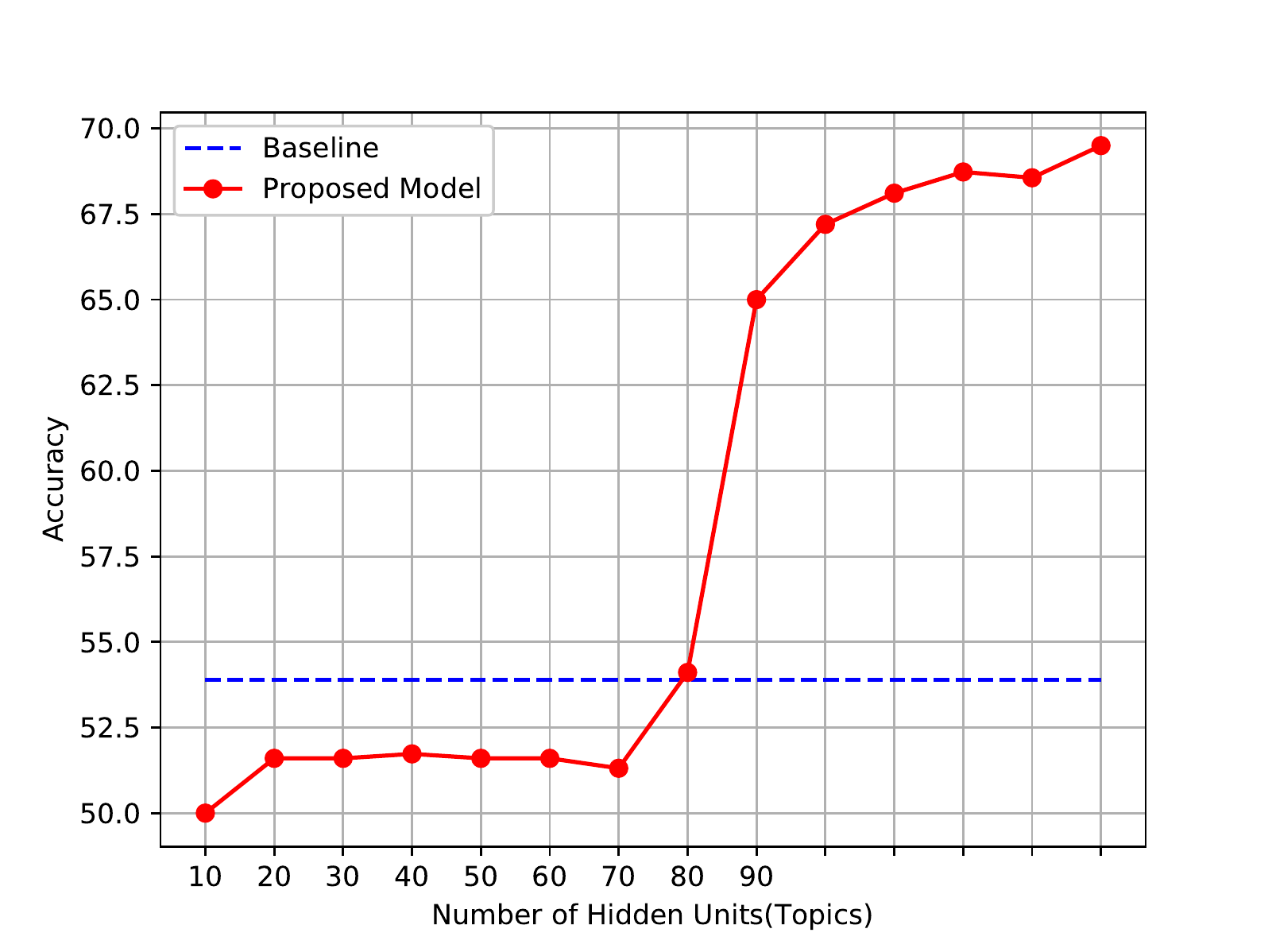}}
	\caption{Sentiment Classification in Movie Review Dataset with Proposed Model and Base Model for Different Number of Topics }
	\label{fig7}
\end{figure}

The sentiment classification results obtained by the proposed approach and the basic model for two different dataset are presented in Fig. \ref{fig7}. To calculate the sentiment classification precision of the proposed model, we used the model trained for 1000 iterations with each dataset and different number of topics.

According to Figs. \ref{fig7-1} and \ref{fig7-2}, in both states, as the number of topics increases, so does the classification precision of the proposed model, and the extent to which it outperforms the basic model.

\begin{figure}[!t]
\centering
\subfigure[Dictionary Size 2000]{\label{fig8-1}\includegraphics[width=70mm]{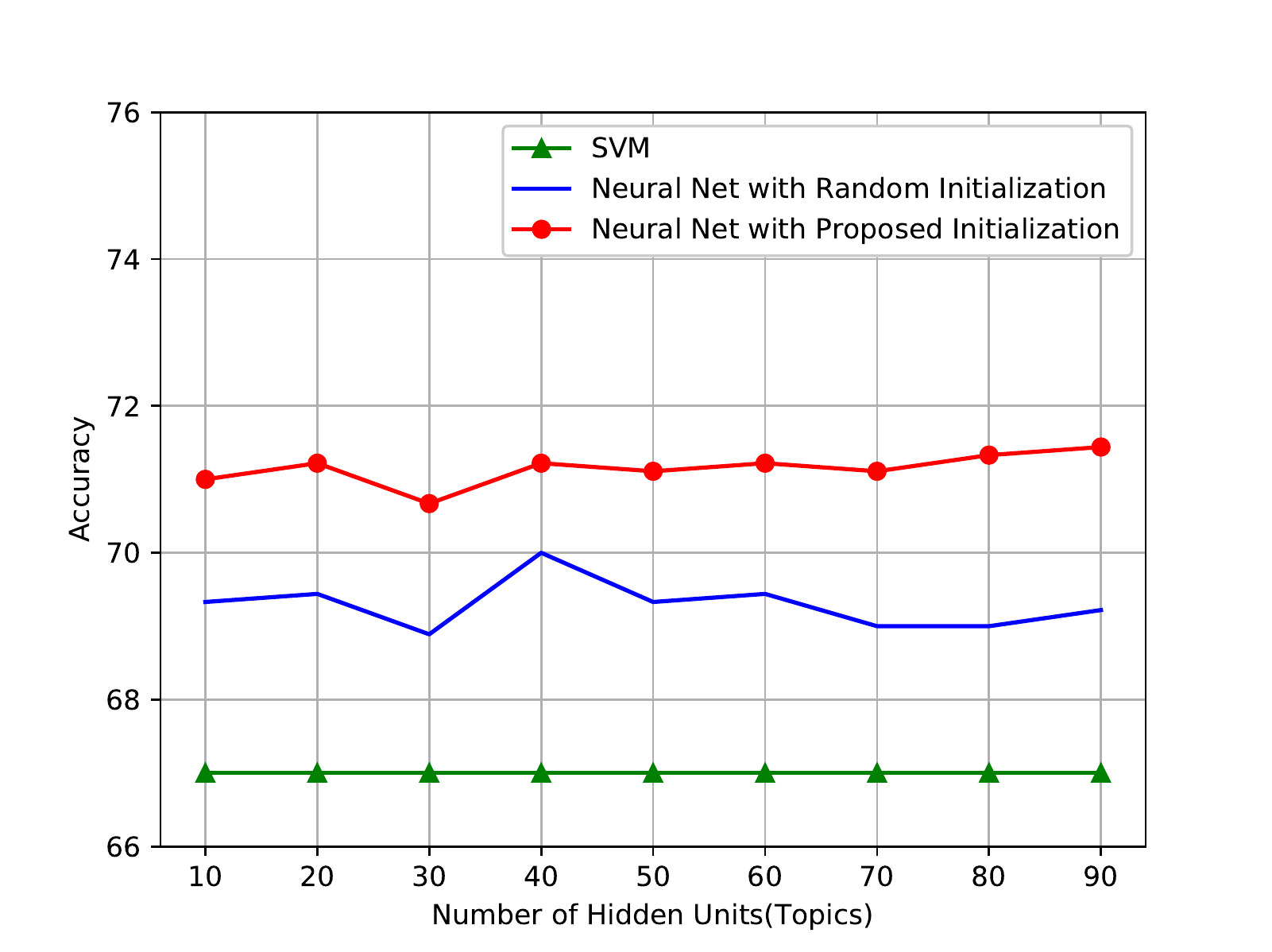}}
\subfigure[Dictionary Size 10000]{\label{fig8-2}\includegraphics[width=70mm]{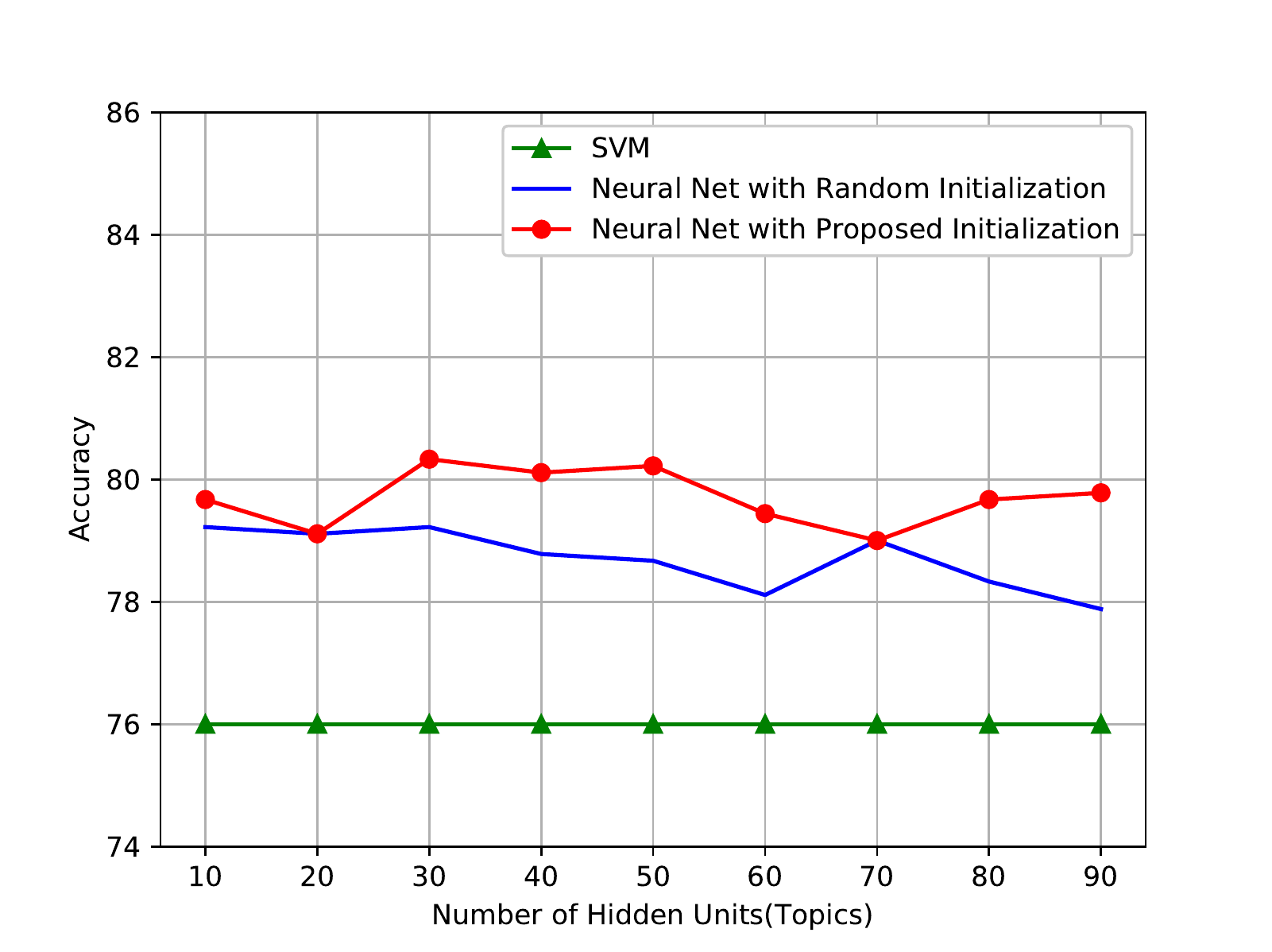}}
\caption{Sentiment Classification in Movie Review Dataset}
\label{fig8}
\end{figure}

Figure \ref{fig8} shows the precision of the sentiment classification performed by two Neural Networks and SVM classification. We used the parameters (weight matrix and bias) obtained in 1000 iterations of training of the proposed model to initialize one of the neural networks and the other one was initialized at random. According to Fig. \ref{fig8}, in both dataset states, the neural network initialized with the values learned by the proposed method has a better precision than the other two methods.

As shown in Fig. \ref{fig8-2}, when using the 10000-word dictionary, the two networks have the same precision only when the number of topics is either 20 or 70. In other cases, the network with non-random initialization has outperformed all other models. In general, we can conclude that the neural network initialized with the values learned by the proposed method has a better sentiment classification performance than other models.

%------------------------------------------
%Conclusion
%-----------------------------------------
\section{Conclusion}
\label{sec5}
In this paper, we presented a novel neural network-based model for the joint sentiment/topic modeling of text data. Review of literature showed the presence of only two Bayesian models, ASUM and JST, for this joint sentiment/topic modeling, and the features and limitations of these model were briefly discussed. The recent developments in the qualities and use of neural networks and the absence of any neural network-based method in the field of joint sentiment/topic modeling were the factors that encouraged the authors to try this approach for this application.

We proposed a supervised neural network-based approach for the joint sentiment/topic modeling of text data. The proposed approach, which falls in the category of generative probabilistic methods, is an extension of the RS model based on the Restricted Boltzmann Machine (RBM) neural network. In the proposed approach, the model is equipped with an additional layer of polynomial probability distribution nature to enable the hidden layer to learn better and more distinct features for each document. This model was trained using a gradient approximation method known as the Contrastive Divergence algorithm.

The proposed model was evaluated using the movie review dataset, the 20-newsgroups dataset, and the multi-domain sentiment dataset, which are the prominent databases for the performance evaluation of topic and sentiment models of text data. We also used perplexity, which is a well-known criterion for the evaluation of generative models, to evaluate the performance of the proposed method in the text data modeling. According to the results, we can claim that incorporating the sentiment into the document modeling, as we did in the present work, will lead to development of generative models of higher quality for document modeling. We also evaluated the data retrieval performance of the proposed method through comparison with the RS model. The results of the tests performed on two databases demonstrated the superior performance and precision of the proposed method in data retrieval from text documents.

\end{document}